# Handling Concept Drifts in Regression Problems – the Error Intersection Approach

Lucas Baier[1], Marcel Hofmann[2], Niklas Kühl[1], Marisa Mohr[2,3] and Gerhard Satzger[1]

[1] Karlsruhe Institute of Technology, Institute of Information Systems and Marketing, Karlsruhe, Germany; [2] inovex GmbH, Karlsruhe, Germany; [3]University of Lübeck, Institute of Information Systems, Lübeck, Germany
`{lucas.baier, niklas.kuehl, gerhard.satzger}@kit.edu,`
`{marcel.hofmann, marisa.mohr}@inovex.de`

**Abstract.** Machine learning models are omnipresent for predictions on big data. One challenge of deployed models is the change of the data over time—a phenomenon called concept drift. If not handled correctly, a concept drift can lead to significant mispredictions. We explore a novel approach for concept drift handling, which depicts a strategy to switch between the application of simple and complex machine learning models for regression tasks. We assume that the approach plays out the individual strengths of each model, switching to the simpler model if a drift occurs and switching back to the complex model for typical situations. We instantiate the approach on a real-world data set of taxi demand in New York City, which is prone to multiple drifts, e.g. the weather phenomena of blizzards, resulting in a sudden decrease of taxi demand. We are able to show that our suggested approach outperforms all regarded baselines significantly.

## 1 Introduction and Related Work

Due to the large increase of data in the last decade, various industries are examining how to reap the benefits of this new resource. Machine Learning (ML) is playing an important role in this context by transforming and (semi-)automating established business processes, spanning from marketing to operations [1]. Typically, companies rely on ML models for increasing the efficiency of their processes or for offering new or improved services and products [2]. Typical applications of ML range from computer vision over speech recognition to natural language processing but also the control of manufacturing robots. Thereby, these techniques are especially influencing data-intensive tasks such as consumer services or the analysis and handling of faults in complex production systems [3]. Nowadays, most of these problems are tackled with supervised ML algorithms [3] where the algorithm depends on labeled training data.

ML can create ongoing value when the resulting models are deployed in the information systems of the respective company and deliver ongoing recommendations and optimized decisions on continuous data streams [4]. However, data streams usually evolve over time and thus, their underlying probability distribution or their data structure changes [5]. The challenge of changing data streams for supervised ML tasks has been described with the term "concept drift" [6]. The joint probability distribution of a set of input variables *X* and the label *y* is described as concept *p(X,y)*. However, "in the real world concepts are not stable but change with time" [7, p.1]. This fact indicates that machine learning models built on previous data might not be suitable for making predictions on new data. Therefore, it is necessary to frequently adapt the prediction approach. A mathematical definition of concept drift can be expressed as follows [8]:

$$\exists X: p_{t0}(X, y) \neq p_{t1}(X, y)$$

This definition explains concept drift as the change in the joint probability distribution between two time points $t_0$ and $t_1$. Changes in the incoming data stream can depend on a multitude of different internal or external influences. Usually, it is impossible to measure all of those possible confounding factors in an environment—which is why this information cannot be included in the predictive features of a ML model. Those factors are considered as "hidden context" of the ML model [7]. Concept drifts in data streams are usually classified into the following types [9]: Sudden or abrupt concept drift refers to situations where the data changes very quickly. A typical example for this drift type is the sudden failure of a sensor. Incremental and gradual concept drift is characterized by slower and more gradual changes, for instance preference shifts in a whole population. Reoccurring drift is determined by seasonal patterns such as ice cream sales in summer. There exists also a more detailed taxonomy for characterizing drifts which also contains categories such as drift duration and magnitude [10].

Figure 1 gives an overview on strategies which can be applied for detecting and handling concept drift. The first dimension refers to the application of an explicit drift detection algorithm. The second dimension describes the adaptations of the underlying ML model. The simplest option is the development of a robust, static ML model which is trained once and then deployed for an ongoing prediction [11], the upper left case in Table 1. Other approaches continuously adapt the prediction model, e.g. with a sliding window where new data instances are continuously used to adapt the prediction model [12]. Such approaches rely on an ongoing adaptation of the prediction model. Depending on the complexity of the model, this requires a lot of computational power. Furthermore, time constraints might also not allow the retraining of the entire model before the next prediction is required, especially in environments with limited resources, e.g. on mobile devices [13]. The lower part of the table depicts approaches which rely on a dedicated drift detection. Drift detection can be handled by an algorithm which detects drifts in the incoming data or the distribution of the prediction error. Based on detected drifts, the model can either be retrained [14] or another model can be applied. Approaches with and without drift detection are also named as active and passive approaches [15]. Various explicit drift detection (active approach) approaches have been proposed, among others the most popular ones such as Page-Hinkley [16], ADWIN [17], EDDM [18]. Those drift detection approaches are often used as benchmarks for new drift detection methods. All of these methods have in common that they observe the misclassification error to detect drifts in the data.

**Model Adaptation**

|  |  | No | Yes |
|---|---|---|---|
| **Drift Detection** | No | Static model (*this work*) | Window-based approach |
|  | Yes | Drift detection with model change (*this work*) | Drift detection with model adaptation |

**Figure 1.** Model Adaptation and Drift Detection Options

Interestingly, predominant approaches for concept drift adaptation focus on *classification* tasks [19] and require the statistical properties of a target variable with binomial distribution. However, many ML challenges need to be modelled as *regression* tasks, e.g. 20 out of 89 studies applying ML and being published in ECIS and ICIS between 2010-2018 are regression problems. In this case, approaches for classification cannot be applied or at least require costly adaptation measures which might potentially harm their performance. Therefore, this work focuses on the application of concept drift strategies for regression tasks which leads to the general research question of this work and the overall research endeavor.

General RQ: *How can we address concept drifts in regression problems?*

Existing approaches for drift detection on regression problems focus on the computation of dedicated drift detection features on the input data in order to detect drifts [19]. In contrast to this, we want to develop an approach based on the prediction error of various models in regression problems. In statistics, a similar problem is the detection of structural changes in time series data [20, 21], a powerful tool to understand and analyze complex interdependencies such as in econometric models [22]. Research in this domain is closely related to unit root testing for time series where the characteristics of a stochastic component (besides a deterministic component) are examined. However, researchers have shown that unit root tests can lead to misleading results when not considering structural breaks in the time series [23]. An application of those methods requires the full input data, i.e. the complete time series, as well as a prior definition of the number of structural breaks to be expected [24]. Therefore, those methods can only be applied in hindsight after the time series has been completed which makes them less suitable for the application in the scenario depicted in this work. An adaptation of a prediction model months or even years after the occurrence of a concept drift does not promise large increases in predictive performance.

Other techniques rely on ensemble methods which have been widely studied and applied for concept drift [25, 26]. Those methods usually rely on an incremental update of each model's importance and parameters. The importance of one model for the overall prediction is decreased and its parameters are adapted if the prediction error of the last prediction is rather large [27].

The novel approach introduced in this paper—labeled as "error intersection approach" (*EIA*)—utilizes static prediction models which are alternated based on the development of the error curve. Static models have the advantage that they need to be implemented only once and can also be scrutinized and tested extensively before they are deployed in production for ongoing predictions. Usually, companies are reluctant to deploy models that adapt and change automatically such as the above described ensemble methods due to the fear of bugs and unexpected behavior [28]. In general, such black-box approaches are regarded critically due to the limited explainability of the issued predictions. Furthermore, our static model approach compared to dynamic models does not need to be retrained frequently which saves a significant amount of computational power as well as time. This advantage is especially important when ML models are deployed on local computing units, such as wireless sensor networks [29].

*EIA* is inspired by the paired learner method for concept drift in classification tasks [30]. This method uses differences in prediction accuracy between a stable—but more accurate ML model and a reactive, simple model to detect drift and to trigger a retraining of the stable model. However, we focus on regression problems and we also do not want to replace existing models in case of drift:

*RQ1: How can we design drift detectors utilizing multiple static models for regression problems?*

For answering this question, we are building *EIA* based on two prediction models, one simple forecast model and a complex neural network model. *EIA* analyzes and takes advantage of the different degrees of complexity between the two models.

As application domain, this work performs demand forecast for mobility solutions which has been investigated before in IS and related disciplines, e.g. by predicting demand for carsharing services [31]. However, this work focuses on the prediction of taxi demand in different taxi zones in New York City (NYC). The dataset is publicly available and provides information about every taxi trip which has been performed since January 2009. Due to the long timespan of the dataset, different types of drift can be observed, which indicates its suitability for the task at hand. Related work already investigated the problem of predicting taxi demand based on this dataset with complex prediction models such as LSTM or convolutional neural networks [32, 33]. However, those approaches focus on optimizing the prediction error on shorter time spans. In contrast, we use this dataset for evaluating the long-term prediction of taxi demand on a test set of 6.5 years under the investigation of concept drift—and do so with our proposed approach.

The remainder of this paper is structured as follows: Section 2 presents the application domain and illustrates some of the existing drifts in the taxi demand data. Section 3 describes the design of *EIA* and the corresponding benchmarks. Section 4 introduces the results and explains the evaluation of our proposed approach. The fifth and the final section discusses our results, describes theoretical and managerial implications, acknowledges limitations and outlines necessary next steps.

## 2 Use Case

This section describes the underlying dataset with taxi rides in NYC as well as some exemplary concept drifts which largely influence the prediction models.

### 2.1 New York City Taxi Dataset

The NYC taxi trip dataset [34] is provided by the New York Taxi and Limousine Commission (TLC) and contains information about all taxi trips that are conducted in NYC. We work with the taxi data from January 2009 up to June 2018. TLC provides information about the taxi trips separately for yellow taxis, green taxis and for-hire-vehicles (FHVs) respectively. Yellow taxis mainly operate within Manhattan, whereas green taxis are only allowed to operate outside of Manhattan. Furthermore, FHVs include ride-hailing services such as Uber. In this work, we are focusing on the yellow taxis since only their data is available for the overall timespan from the beginning in 2009. By focusing on a long-term duration, we expect more frequent and more significant concept drifts (e.g. weather, rise of Uber) to be present. In total, this gives us access to around 1.4 billion rides with yellow cabs.

With regard to preprocessing the data, we first remove outliers where distance or duration of a taxi ride are equal to zero. All trips before 2016 contain the exact geolocation of the start as well as the end of the taxi trip. All subsequent taxi trips only refer to the more high-level taxi-zones of pickup and drop-off of the passengers. Therefore, we match the previous exact geolocation data with the taxi zones introduced in 2016. Subsequently, we aggregate all taxi trips to identify the hourly demand per taxi zone. In this work, we focus on the 20 largest taxi zones because those already account for 60% of the overall taxi demand. This leads to a demand history with 83,231 hourly taxi demands for each of the 20 taxi zones.

### 2.2 Exemplary Drifts

To lay the foundation for our work, we describe exemplary concept drifts which we have identified in the taxi demand dataset. One source of change is the market entry of new competitors in the passenger carriage business which has already been discussed in related literature [35]. Uber already launched its service in NYC in 2011 with a small fleet of drivers. However, the tracking of FHVs by the TLC only started back in 2015. Therefore, we do not have any information with regard to the use of Uber, Lyft etc. before that date. Figure 2 shows the overall demand trajectories for both Yellow cabs and Uber over the entire time span. At first, demand for yellow cabs rises steadily between 2009 and 2012. After 2012, however, the overall trend clearly indicates a decreasing demand due to the rise of new competitors. The typical demand pattern during the course of a year stays fairly constant during the whole duration. Referring to the previously introduced concept drift patterns, the decreasing demand for yellow taxis over time can be described as an incremental concept drift where data patterns slowly evolve. However, we need to be careful with assumptions about the exact time as well as impact of drifts in this real-world dataset since there is no ground truth describing the exact characteristics of this drift as opposed to simulated data [7]. Furthermore, changes in the real-world are often related to a multitude of factors.

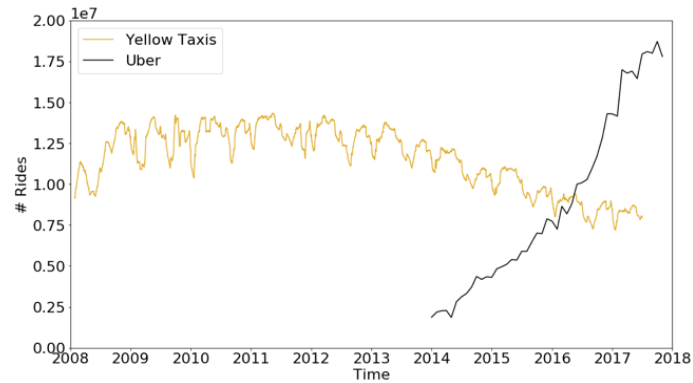

**Figure 2.** Overall NYC yellow cab and Uber demand per month

Another source of drift in the dataset are extreme weather events such as hurricanes or thunderstorms. Taxi demand naturally adapts to those unusual weather situations. Figure 3 depicts the taxi demand during the course of Tuesday, January 27$^{th}$, 2015 (in blue) and the average demand on Tuesdays in 2015 (in red) as well as the 25% and 75%-quantile of the average demand.

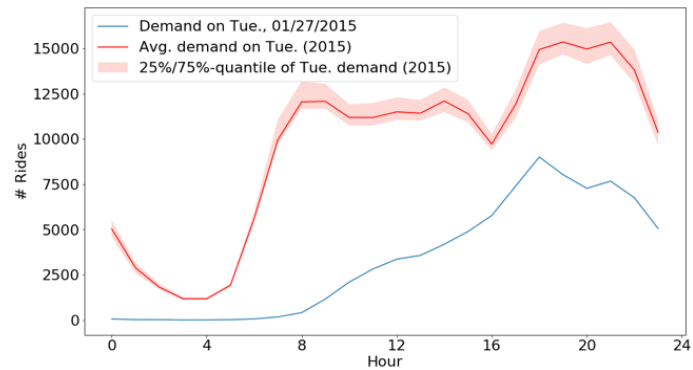

**Figure 3.** Taxi demand during a blizzard on 2015-01-27

It is obvious that the average demand and the demand on January 27$^{th}$ clearly deviate. The blue line indicates nearly zero demand during nighttime and early morning which is due to a blizzard which passed by NYC with declared snow emergencies as well as enacted travel bans. In contrast to the incremental drift example above, this event can be regarded as a sudden concept drift. Another example for sudden drifts in the dataset is the occurrence of special events such as festivals in dedicated taxi zones. In this case, the demand for taxi rides is suddenly increased dramatically in comparison to usual demand patterns.

When applying a ML model to predict future taxi demand, we are aware that one could probably increase significantly the predictive power of a model by including external data such as weather data, competitor data or an events calendar. However, the focus of this work is not to provide the best possible demand prediction. We aim at examining and quantifying the effect of concept drifts in real-world situations. Therefore, we will consider weather and other facts as external hidden variables (see Section 1) that we cannot observe in the application environment. This requires an adequate preparation and adaptation of the applied prediction model.

In this particular use case, it seems rather easy to identify factors (e.g. weather etc.) which have a large influence on the prediction power of a model as well as how to include this information as predictive features. This might also be due to the nature of the overall project since nearly everyone has already used a taxi as a means of transportation. However, in hindsight, it is often easier to identify unusual demand patterns and subsequently investigate the underlying reason. For a predictive model, though, this information is required in real time. Furthermore, including weather features in this use case and disregarding other unidentified influencing factors might lead to overfitting of drift behavior on weather phenomena.

In other use cases and application areas, it is often very difficult or impossible at all to identify influencing variables apart from the obvious dataset [36]. In case those can be identified, it is often impossible to measure and quantify those factors. Therefore, we decide to restrain the inclusion of external data sources in this work. Furthermore, after a thorough analysis of our data, we also have identified a lot of fluctuations and abnormalities in the dataset where it is impossible to identify the underlying reason without additional knowledge. Usually, drift detection and adaption approaches are evaluated based on simulated datasets. In this work, we want to evaluate our drift detection approach (*EIA*) based on a real-world dataset.

## 3 Design of the Error Intersection Approach

With the introduction of various examples for drifts in the NYC taxi dataset at hand, it is necessary to develop a prediction strategy which accounts for those concept drifts and provides reasonable predictions. As described in Section 2, we assume incremental as well as sudden concept drift to be present in this dataset. Since those two types of concept drifts are fundamentally different and require adapted handling strategies each, we decide to focus on sudden concept drift in this work. We propose an approach which relies on two different prediction models.

Ensemble methods have been widely used in concept drift adaptation methods [8]. Usually those approaches rely on the combination of various models with an average of the delivered predictions to increase the overall performance. However, in this work, we propose a different approach which uses the predictions issued by two static models with different complexity to detect drift in the data and adapt the prediction accordingly. This approach is depicted in figure 4. As first model, we use a simple model ($M_{simple}$) which is only influenced by the most recent demand in the respective taxi zone. As second model, we apply a neural network ($M_{complex}$) which receives as input a large demand history over all taxi zones. During normal times, $M_{complex}$ is applied because it successfully captures the general demand structure and therefore is able to compute accurate predictions for the taxi demand in the respective taxi zone. However, during times with sudden concept drifts, $M_{complex}$ cannot provide accurate predictions since the demand patterns clearly deviate from the usual trajectories. In those cases, $M_{simple}$ is applied because it can quickly adapt to current demand changes. By design, this approach is presumably only able to deal with sudden concept drift since incremental drifts require frequent adaptations of the predictions models which is not the focus of this work. The switch between models is triggered by an intersection of the prediction error curves of $M_{complex}$ and $M_{simple}$. Therefore, we call this the "error intersection approach" (*EIA*). *EIA* is feasible since we always receive the true label for the prediction after the course of one hour.

*EIA* is inspired by a streaming architecture for deploying ML models [28] which suggests the deployment of several individual and independent prediction models. This way, it can be guaranteed that a prediction can always be issued in time for a new data instance. Furthermore, *EIA* is based on the paired learner approach [30] which uses differences in prediction accuracy between a stable and a reactive ML model to detect drifts.

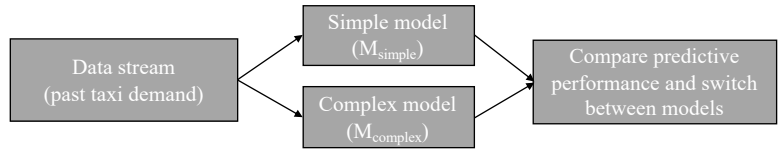

**Figure 4.** Design approach for the prediction

$M_{simple}$ is a baseline model often used in time series forecasting literature [37]. It just predicts the demand value from the last period in the respective taxi zone. This model does not learn any parameters but is very good at capturing current trends.

$M_{complex}$ is a neural network which contains as input features the regions and the current weekday as one-hot encoded vector. Furthermore, it receives the demand of the last 24 hours as well as the demand during the same hour on the same weekday in the four past weeks. Additionally, we include cosine and sine features to depict that hours and months are cyclical features to improve prediction performance as suggested in literature [38].

$$cos_{hour} = \cos\left(\frac{h * 2 * \pi}{24}\right); \; sin_{hour} = \sin\left(\frac{h * 2 * \pi}{24}\right)$$

To compute the respective features regarding months, the denominator is adapted to 12. We use 128 neurons in the hidden layer with a 'relu' activation function and the network is trained using 50% dropout. Similar network architectures have been used before for taxi demand prediction [39].

## 4 First Evaluation

This section introduces first results with the previously proposed design. Figure 5 illustrates the applied combinations of drift detectors and models for the prediction of the taxi demand. The upper part of the table (in red) describes the combinations that have been implemented so far. The lower part of the table (in blue) contains the options that need to be pursued in future work.

|  | Ensemble | Drift Detector | Applied prediction model | Retraining after drift detection |
|---|---|---|---|---|
| This work | No | n/a | 1 ($M_{simple}$ *or* $M_{complex}$) | No |
| This work | Yes | n/a | 2 ($M_{simple}$ *and* $M_{complex}$) | No |
| This work | No | Page-Hinkley ADWIN EDDM **EIA** | 2 ($M_{simple}$ *and* $M_{complex}$) | No |
| Future work | No | **EIA** | N models (e.g. LSTM) | No |
| Future work | No | **EIA** | N models (e.g. LSTM) | Yes |

**Figure 5.** Overview of applied drift-detector and model combinations in this work and for future work

As error measure, we apply the root mean squared error (RMSE) which is the standard metric to evaluate taxi demand predictions on the NYC dataset (e.g., [33]). Furthermore, we compute the symmetric mean absolute percentage error (SMAPE) as a relative error measure. Demand from 2009 up to 2011 is considered as training data, whereas demand after 2012 is used as test data.

As baseline, the performance of $M_{complex}$ and $M_{simple}$ alone on the dataset is evaluated. $M_{simple}$ does not contain any parameters and therefore cannot be updated. However, with regard to $M_{complex}$, we retrain the weights once a year so that $M_{complex}$ can adapt to the general trend of the taxi demand over the years. This means that the forecast for 2012 is performed with a model trained on data from 2009-2011, the forecast for 2013 is issued by a model trained on data from 2010-2012. As additional baseline, we build an ensemble from both models' predictions since existing drift handling strategies for regression usually are built this way (see Section 1). We compute the exponential weighted moving average (EWMA) of the last 6 predictions errors of both models respectively and determine the sum of errors. Subsequently, we compute the contribution of each model to the sum of errors to determine the weights of each model for the ensemble prediction (e.g. if $M_{complex}$ accounts for 1/3 of the sum of errors, its weight for the next prediction are 2/3). Furthermore, we test the established methods Page-Hinkley (PH), ADWIN and EDDM as drift detectors. When those methods detect a drift, the switch between the two prediction models is performed. Since EDDM can only be applied to classification problems, we need to transform the regression problem [25]. *EIA* (see Section 3), in contrast, switches between models based on the EWMA of the prediction errors in the last 6 hours. The model with the lower recent prediction error (either $M_{simple}$ or $M_{complex}$) is the active model for computing the next prediction.

After issuing the prediction, the error terms are evaluated once more and the model for the subsequent hourly prediction is selected.

Table 2 introduces the results for the overall prediction performance of the different approaches on the test set. Not surprisingly, $M_{simple}$ produces the highest RMSE by far, which portrays the worst result. This model is just too simple for producing a good overall forecast. In contrast, $M_{complex}$ already performs well with an RMSE of 50.478. Standard drift detection methods seem to work reasonably on this dataset; however, their application leads to a worse performance compared to $M_{complex}$. *EIA* is depicted in the last row and shows a better performance than $M_{complex}$ alone. The amount of model switches is depicted in the second column.

**Table 1.** First results of EIA in comparison to benchmarks, based on RMSE and SMAPE (the lower the better)

| *Approach* | *# model switches* | *RMSE* | *SMAPE* |
|---|---|---|---|
| $M_{simple}$ | n/a | 115.871 | 13.80% |
| $M_{complex}$ | n/a | 50.478 | 6.01% |
| Ensemble (EWMA) | n/a | 58.381 | 6.75% |
| PH | 1,811 | 82.176 | 9.79% |
| ADWIN | 70 | 97.657 | 11.64% |
| EDDM | 30 | 112.783 | 13.47% |
| **EIA** | **365** | **50.370** | **5.98%** |

The effectiveness of *EIA* is illustrated in figure 6 which depicts the demand predictions during the blizzard in 2015 (see figure 2). In the beginning, *EIA* (red dashed line) always chooses $M_{complex}$ because of the lower prediction error (black line). However, at around 16:00 of January 26[th] (marked by a black vertical line), the approach switches to $M_{simple}$ (lower error curve in grey), which can quickly adapt to the unusual demand pattern. $M_{complex}$ clearly fails to predict this behavior correctly (e.g., peak at around 05:00).

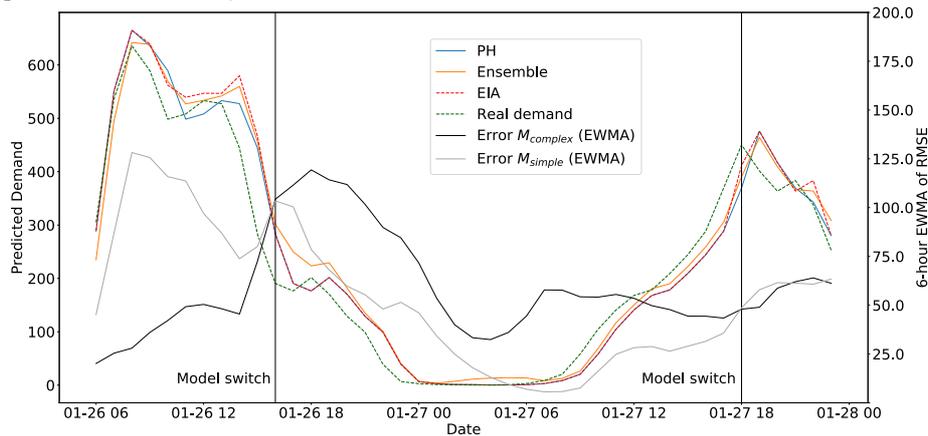

**Figure 6.** Predictions of EIA during blizzard on 2015-01-27

## 5   Discussion

The absolute difference in RMSE between *EIA* and $M_{complex}$ seems to be rather small. However, in total, we average over more than 1.13 million predictions. Therefore, we perform a Diebold-Mariano test to compare the predictive performance between *EIA* and $M_{complex}$ [40]. With a p-value of $1.89 * 10^{-7}$, we can conclude that there is a significant difference in forecast performance between the two approaches. The absolute small difference can be explained by the fact EIA only chooses $M_{simple}$ in 1.24% of all necessary forecasts (706 out of 56,951 hourly forecasts in total). This means that in 98.76% of all forecasts, the predictions of EIA and $M_{complex}$ are the same. However, during those 706 hourly forecasts where $M_{simple}$ is chosen by *EIA,* the prediction performance is largely improved by 8.4% (RMSE *EIA*: 75.31 vs RMSE $M_{complex}$: 82.21). This is a good indication for the effectiveness of the approach and the share of $M_{simple}$ might be larger on a different dataset with corresponding impact on the difference in predictive performance.

The superiority of EIA compared to established drift detectors such as PH can probably be explained by information asymmetry: EIA uses the information of two prediction models and their error curve to select the current optimal model. PH, in contrast, works by analyzing the development of the prediction error of only one prediction model (in our case, either $M_{complex}$ or $M_{simple}$) and therefore has access to less information resulting in model switches at unfavorable points in time.

This argument does not hold true for the ensemble approach since both models are used to compute the resulting predictions. However, performance loss in this case might be explained by the overall bad performance of $M_{simple}$: The weight for each model is determined by considering the past prediction errors which will generally lead to a high weight for $M_{complex}$. Nevertheless, $M_{simple}$ will almost always also receive a weight larger than zero, thereby negatively impacting the ensemble prediction.

Furthermore, we analyze for *EIA* on which days $M_{simple}$ is applied the most. This way, we can identify in hindsight the days with most significant concept drifts. Table 2 shows an excerpt of days with a frequent use of $M_{simple}$ for a prediction as well as the corresponding special events (drift cause) on that day. The second column depicts the absolute improvement in RMSE of *EIA* compared to predictions by $M_{complex}$ alone. In most cases, drift is triggered by unusual weather events or public holidays. However, we also find several days (e.g. August 1st, 2013) where we are not able to identify the underlying reason for the drift cause. This depicts the strengths of *EIA* since it does not require any additional data compared to an explicit integration of features such as weather or public holidays (see explanation in Section 2). Furthermore, we compared taxi demand during New Year's Eve for several years: In 2012, for instance, taxi demand peaked before midnight whereas in 2017, the highest demand occurred after midnight. This variability complicates the learning process even if an explicit feature for holidays is included. In future work, we want to perform a more comprehensive analysis of these results in order to understand when it is most suitable to apply the approach.

Table 2. Excerpt of days with frequent use of $M_{simple}$ for a prediction

| Date | RMSE Improvement (abs.) | # Predictions by $M_{simple}$ | Probable Drift Cause |
|---|---|---|---|
| 2012-07-04 | 5.07 | 14/24 | 4$^{th}$ of July |
| 2012-10-29 | 24.41 | 22/24 | Hurricane Sandy |
| 2012-12-25 | 9.22 | 17/24 | Christmas Day |
| 2013-08-01 | 9.35 | 10/24 | ? (unknown) |
| … | … | … | … |
| 2017-06-25 | 5.48 | 10/24 | ? (unknown) |
| 2018-03-21 | 15.21 | 14/24 | Cyclone (Nor'easter) |

## 6   Conclusion

In the work at hand, we explore a novel approach ("error intersection approach") for concept drift handling for supervised regression tasks. Established drift detection methods usually focus on classification problems. Our approach, in its core, depicts a strategy to switch between the application of simple and complex prediction models which is designed to deliver superior performance results in real-world data sets prone to concepts drifts. We hypothesize that the drift detector allows to play out the individual strengths of each model, switching to the simpler model if a drift occurs and switching back to the complex model for typical situations. To illustrate our suggestion, we instantiate the approach on a real-world data set of taxi demand in NYC. For this very data set, we are aware of multiple drifts, e.g. short-term drifts such as the weather phenomena of a blizzard. We apply different, typical predictive models for regression tasks and are able to show that our suggestion outperforms all regarded baselines significantly.

Obviously, these results are preliminary and have certain limitations. Our prediction is presumably worse than very complex CNN and LSTM architectures [32, 39]. Also, we have not tested other powerful machine learning techniques such as XGBoost [41]. However, previous work has evaluated those models only on shorter test periods (2 and 6 months respectively). Furthermore, models from related work reveal no insights on their effectiveness for drift handling, while *EIA* offers more transparency (e.g., how often were the model switched, when was which model used, etc.) and, therefore, allows for more interpretability [42]. Still, the applied approach is (presumably) only meaningful when sudden concept drift is expected. To further explore this, more research is required to formulate clear guidelines on the precise cases in which we can recommend the use of the suggested approach. Furthermore, our approach is only feasible when the true label for a delivered prediction can be acquired afterwards, which might not be the case in all applications. However, this limitation also holds true for established drift detection methods.

In future work, we aim to further develop *EIA* regarding several aspects. To examine generalizability, we aim to test the effectiveness of *EIA* on a different dataset. A simulation study might be a worthwhile tool in this context. Furthermore, we want to extend our work on the drift detection algorithm. Additionally, on the presented data set, it will be interesting to identify regions with the highest drifts where it is most appropriate to apply *EIA*. Furthermore, *EIA* in its current form is rather basic, as we only regard one change detection and only switch between two models—a simple and a complex one. In future work, we aim to employ more sophisticated change algorithms, but also investigate approaches with more models, e.g. very simple/average/very complex. Finally, we did not regard models with immediate retraining after the drift detection—which also remains an interesting option for future work.

## References

1. Chen, H., Chiang, R.H.L., Storey, V.C.: Business Intelligence and Analytics: From Big Data To Big Impact. Mis Q. 36, 1165–1188 (2012).
2. Schüritz, R., Satzger, G.: Patterns of Data-Infused Business Model Innovation. In: CBI 2016. pp. 133–142 (2016).
3. Jordan, M.I., Mitchell, T.M.: Machine learning: Trends, perspectives, and prospects, (2015).
4. Baier, L., Kühl, N., Satzger, G.: How to Cope with Change? Preserving Validity of Predictive Services over Time. In: Hawaii International Conference on System Sciences (HICSS-52) (2019).
5. Aggarwal, C.C., Watson, T.J., Ctr, R., Han, J., Wang, J., Yu, P.S.: A Framework for Clustering Evolving Data Streams. Proc. - 29th int. conf. very large data bases. 81–92 (2003).
6. Widmer, G., Kubat, M.: Learning in the presence of concept drift and hidden contexts. Mach. Learn. 23, 69–101 (1996).
7. Tsymbal, A.: The problem of concept drift: definitions and related work. Comput. Sci. Dep. Trinity Coll. Dublin. 4, 2004–15 (2004).
8. Gama, J., Žliobaitė, I., Bifet, A., Pechenizkiy, M., Bouchachia, A.: A survey on concept drift adaptation. ACM Comput. Surv. 46, 1–37 (2014).
9. Zliobaite, I.: Learning Under Concept Drift: An Overview. arXiv Prepr. (2010).
10. Webb, G.I., Hyde, R., Cao, H., Nguyen, H.L., Petitjean, F.: Characterizing concept drift. Data Min. Knowl. Discov. 30, 964–994 (2016).
11. Guajardo, J.A., Weber, R., Miranda, J.: A model updating strategy for predicting time series with seasonal patterns. Appl. Soft Comput. J. (2010).
12. Kuncheva, L., Žliobaite, I.: On the window size for classification in changing environments. Intell. Data Anal. 13, 861–872 (2009).
13. Oneto, L., Ghio, A., Ridella, S., Anguita, D.: Learning Resource-Aware Classifiers for Mobile Devices: From Regularization to Energy Efficiency. Neurocomputing. (2015).
14. Ivannikov, A., Pechenizkiy, M., Bakker, J., Leino, T., Jegoroff, M., Kärkkäinen, T., Äyrämö, S.: Online mass flow prediction in CFB boilers. In: Lecture Notes in Computer Science. pp. 206–219 (2009).
15. Ditzler, G., Roveri, M., Alippi, C., Polikar, R.: Learning in Nonstationary Environments: A Survey. IEEE Comput. Intell. Mag. 10, 12–25 (2015).
16. Page, E.S.: Continuous inspection schemes. Biometrika. 41, 100–115 (1954).
17. Bifet, A., Gavalda, R.: Learning from time-changing data with adaptive windowing. In: Proceedings of the 2007 SIAM international conference on data mining. pp. 443–448 (2007).

18. Baena-Garcia, M., del Campo-Ávila, J., Fidalgo, R., Bifet, A., Gavalda, R., Morales-Bueno, R.: Early drift detection method. In: Fourth international workshop on knowledge discovery from data streams. pp. 77–86 (2006).
19. Cavalcante, R.C., Minku, L.L., Oliveira, A.L.I.: FEDD: Feature Extraction for Explicit Concept Drift Detection in time series. Proc. Int. Jt. Conf. Neural Networks. 2016-Octob, 740–747 (2016).
20. Zeileis, A., Kleiber, C., Krämer, W., Hornik, K.: Testing and dating of structural changes in practice. Comput. Stat. Data Anal. 44, 109–123 (2003).
21. Verbesselt, J., Hyndman, R., Newnham, G., Culvenor, D.: Detecting trend and seasonal changes in satellite image time series. Remote Sens. Environ. 114, 106–115 (2010).
22. Fernald, J.G., Hall, R.E., Stock, J.H., Watson, M.W.: The disappointing recovery of output after 2009. (2017).
23. Perron, P.: The great crash, the oil price shock, and the unit root hypothesis. Econom. J. Econom. Soc. 1361–1401 (1989).
24. Glynn, J., Perera, N., Verma, R.: Unit Root Tests and Structural Breaks: A Survey with Applications. Rev. Métodos Cuantitativos para la Econ. y la Empres. 3, (2007).
25. Xiao, J., Xiao, Z., Wang, D., Bai, J., Havyarimana, V., Zeng, F.: Short-term traffic volume prediction by ensemble learning in concept drifting environments. Knowledge-Based Syst. 164, 213–225 (2019).
26. Sun, J., Fujita, H., Chen, P., Li, H.: Dynamic financial distress prediction with concept drift based on time weighting combined with Adaboost support vector machine ensemble. Knowledge-Based Syst. 120, 4–14 (2017).
27. Soares, S.G., Araújo, R.: A dynamic and on-line ensemble regression for changing environments. Expert Syst. Appl. 42, 2935–2948 (2015).
28. Dunning, T., Friedman, E.: Machine Learning Logistics. O'Reilly Media, Inc. (2017).
29. Alsheikh, M.A., Lin, S., Niyato, D., Tan, H.P.: Machine learning in wireless sensor networks: Algorithms, strategies, and applications. IEEE Commun. Surv. Tutorials. (2014).
30. Bach, S.H., Maloof, M.A.: Paired learners for concept drift. Proc. - IEEE Int. Conf. Data Mining, ICDM. 23–32 (2008).
31. Kahlen, M., Ketter, W., Lee, T., Gupta, A.: Optimal Prepositioning and Fleet Sizing to Maximize Profits for One-Way Transportation Companies. ICIS Proc. (2017).
32. Xu, J., Rahmatizadeh, R., Boloni, L., Turgut, D.: Real-Time prediction of taxi demand using recurrent neural networks. IEEE Trans. Intell. Transp. Syst. 19, 2572–2581 (2018).
33. Zhang, J., Zheng, Y., Qi, D.: Deep Spatio-Temporal Residual Networks for Citywide Crowd Flows Prediction. (2016).
34. TLC: Taxi and Limousine Commission (TLC) Trip Record Data, https://www1.nyc.gov/site/tlc/about/tlc-trip-record-data.page.
35. Cramer, J., Krueger, A.B.: Disruptive change in the taxi business: The case of Uber. Am. Econ. Rev. 106, 177–182 (2016).
36. Stowers, K., Kasdaglis, N., Newton, O., Lakhmani, S., Wohleber, R., Chen, J.: Intelligent agent transparency: The design and evaluation of an interface to facilitate human and intelligent agent collaboration. In: Proceedings of the Human Factors and Ergonomics Society Annual Meeting. pp. 1706–1710 (2016).
37. Hyndman, R.J., Athanasopoulos, G.: Forecasting: principles and practice. OTexts (2018).
38. Hernández, L., Baladron, C., Aguiar, J.M., Carro, B., Sanchez-Esguevillas, A., Lloret, J.,

Chinarro, D., Gomez-Sanz, J.J., Cook, D.: A multi-agent system architecture for smart grid management and forecasting of energy demand in virtual power plants. IEEE Commun. Mag. 51, 106–113 (2013).
39. Liao, S., Zhou, L., Di, X., Yuan, B., Xiong, J.: Large-scale short-term urban taxi demand forecasting using deep learning. In: 2018 23rd Asia and South Pacific Design Automation Conference (ASP-DAC). pp. 428–433. IEEE (2018).
40. Diebold, F.X., Mariano, R.S.: Comparing predictive accuracy. J. Bus. Econ. Stat. 20, 134–144 (2002).
41. Chen, T., Guestrin, C.: Xgboost: A scalable tree boosting system. In: Proceedings of the 22nd acm sigkdd international conference on knowledge discovery and data mining. pp. 785–794 (2016).
42. Gilpin, L.H., Bau, D., Yuan, B.Z., Bajwa, A., Specter, M., Kagal, L.: Explaining Explanations: An Overview of Interpretability of Machine Learning. In: 2018 IEEE 5th International Conference on Data Science and Advanced Analytics (DSAA). pp. 80–89 (2018).